\pdfoutput=1
\documentclass[11pt]{article}
\usepackage{acl2021}

\usepackage[utf8]{inputenc}

\usepackage{times}
\usepackage{latexsym}
\usepackage{xspace}
\usepackage{enumitem}

\usepackage{url}
\usepackage{xcolor}
\usepackage{booktabs}
\usepackage{diagbox}
\usepackage{multirow}
\usepackage{balance}
\usepackage{amsfonts}
\usepackage{amsmath}
\usepackage{amssymb}
\usepackage{graphicx}
\usepackage{subfigure}
\usepackage{microtype}
\usepackage{wrapfig,lipsum}
\usepackage{algorithm}
\usepackage{algorithmicx}
\usepackage{algpseudocode}
\usepackage{makecell}
\usepackage[T1]{fontenc}
\usepackage{soul}
\usepackage{tabularx, booktabs}
\usepackage{todonotes}
\usepackage{comment}
\newcommand{\cbb}{\mathbf{c}}
\newcolumntype{Y}{>{\centering\arraybackslash}X}
\newcolumntype{L}{>{\arraybackslash}X}
\newcolumntype{M}[1]{>{\centering\arraybackslash}m{#1}}

\algnewcommand\algorithmicinput{\textbf{Input:}}
\algnewcommand\algorithmicoutput{\textbf{Output:}}
\algnewcommand\algorithmicparameter{\textbf{Parameters:}}
\algnewcommand\INPUT{\item[\algorithmicinput]}
\algnewcommand\OUTPUT{\item[\algorithmicoutput]}
\algnewcommand\PARAMETER{\item[\algorithmicparameter]}

\newcommand{\clstoken}{\texttt{[CLS]}}

\newcommand{\ourmodel}{\texttt{UDT-QA}}

\newcommand{\eat}[1]{\ignorespaces}
\input{Definitions.tex}

\usepackage{todonotes}

\newcommand{\chenghao}[2][]{\todo[color=green,size=\scriptsize,fancyline,caption={},#1]{Hao:#2}}

\newcommand\blfootnote[1]{%
  \begingroup
  \renewcommand\thefootnote{}\footnote{#1}%
  \addtocounter{footnote}{-1}%
  \endgroup
}

\title{
Open Domain Question Answering with A Unified Knowledge Interface
}

\author{
Kaixin Ma\textsuperscript{$\clubsuit$}$\dagger^*$,
Hao Cheng\textsuperscript{$\spadesuit$}$^*$,
Xiaodong Liu\textsuperscript{$\spadesuit$},
Eric Nyberg\textsuperscript{$\clubsuit$},
Jianfeng Gao\textsuperscript{$\spadesuit$}
 \\ 
  \textsuperscript{$\clubsuit$} Carnegie Mellon University
  \textsuperscript{$\spadesuit$} Microsoft Research 
 \\
  {\tt \{kaixinm,ehn\}@cs.cmu.edu}
  {\tt \{chehao,xiaodl,jfgao\}@microsoft.com}
}

\date{}

\begin{document}
\maketitle
\begin{abstract}
The retriever-reader framework is popular for open-domain question answering (ODQA) due to its ability to use explicit knowledge.
Although prior work has sought to increase the knowledge coverage by incorporating structured knowledge beyond text, accessing heterogeneous knowledge sources through a unified interface remains an open question. 
While data-to-text generation has the potential to serve as a universal interface for data and text, its feasibility for downstream tasks remains largely unknown.
In this work, we bridge this gap and use the data-to-text method as a means for encoding structured knowledge for ODQA.
Specifically, we propose a \textit{verbalizer-retriever-reader} framework for ODQA over data and text where verbalized tables from Wikipedia and graphs from Wikidata are used as augmented knowledge sources.
We show that our \textbf{U}nified \textbf{D}ata and \textbf{T}ext \textbf{QA}, \ourmodel,
can effectively benefit from the expanded knowledge index, leading to large gains over text-only baselines.
Notably, our approach sets the single-model state-of-the-art on Natural Questions.
Furthermore, our analyses indicate that verbalized knowledge is preferred for answer reasoning for both adapted and hot-swap settings. 
\blfootnote{$\dagger$ Work done during an internship at Microsoft Research} \blfootnote{$*$ Equal contribution}
\end{abstract}


\section{Introduction}
\label{sec:intro}

Pretrained language models \cite{devlin-etal-2019-bert,gpt3} have been shown to store certain knowledge (linguistic or factual) implicitly in parameters \cite{Manning2020linguistic,petroni-etal-2019-language,roberts-etal-2020-much}, partially explaining the superior generalization abilities over downstream tasks.
However, besides the well-known hallucination issue, the \textit{implicit knowledge} learned through language modeling objective over text struggles at reflecting up-to-date knowledge from text and structured data for answering open-domain questions. 
To overcome this, recent work on open domain question answering (ODQA) focuses on the semi-parametric method \cite{karpukhin-etal-2020-dense,guu2020realm} where the pretrained language models can leverage external \textit{explicit knowledge} sources for reasoning.
For example, in the \textit{retriever-reader} framework \citep[][\textit{inter alia}]{min2021neurips}, the reader produces answers by grounding on the relevant evidence from the retriever, the interface to the explicit knowledge source (Wikipedia text passages).
In this work, we focus on the semi-parametric approach for ODQA going beyond textual knowledge.
Specifically, we are interested in the question: \textit{Can we develop a viable unified interface over a realistic heterogeneous knowledge source containing both data and text?}

Recent retriever-reader models \cite{oguz2020unikqa,agarwal-etal-2021-knowledge} have demonstrated that expanding the textual knowledge source with more structured data is beneficial.
However, only knowledge base (KB) is considered in \cite{agarwal-etal-2021-knowledge}, limiting the applicability of their method to other structured data. 
In \cite{oguz2020unikqa}, both tables and KB triples are simply linearized as inputs to the reader, but different retrievers are required for individual cases. 
Here, we propose a \textit{verbalizer-retriever-reader} semi-parametric framework, \ourmodel,
which provides a unification of both representation and model for ODQA over data and text. The key idea is to augment the retriever with a data-to-text verbalizer for accessing heterogeneous knowledge sources, \ie KB graphs from WikiData, tables and passages from Wikipedia.

Given its potential in providing a universal interface for data and text, data-to-text generation is increasingly popular \cite{gardent-etal-2017-webnlg,parikh-etal-2020-totto,nan-etal-2021-dart} with various methods developed recently for converting structured knowledge into natural language \cite{wang-etal-2020-towards, ribeiro2020investigating, chen-etal-2020-kgpt}.
Nevertheless, most existing work has focused on \textit{intrinsic evaluations} exclusively, i.e. the quality of generated text measured by metrics like BLEU \cite{papineni-etal-2002-bleu}, leaving its usefulness on downstream tasks largely unknown.
Moreover, it remains unclear whether a single data-to-text model is able to verbalize heterogeneous structured data effectively.
To bridge the gap, we develop a novel data-to-text generation paradigm for our framework.
We introduce data filtering and beam selection to maximize the faithful coverage of the input information.
To remedy the lack of in-domain data, we further propose an iterative training approach to augment the existing data-to-text training set with high quality outputs selected from the target domain.
With this verbalizer, we convert all tables from Wikipedia (10x more than \cite{oguz2020unikqa}) and sub-graphs from Wikidata together with Wikipedia text passages as the knowledge source for ODQA.

We first validate our data-to-text method using intrinsic metrics on DART \cite{nan-etal-2021-dart} and additional faithfulness evaluation on the target ODQA data. We show that our data-to-text approach can effectively improve the target-domain faithful metric without compromising too much on the intrinsic metrics.
To further evaluate the end-to-end effectiveness, we experiment with \ourmodel\space on the ODQA task using a recent state-of-the-art (SOTA) retriever-reader pipeline, including DPR \cite{karpukhin-etal-2020-dense} and UnitedQA \cite{cheng-etal-2021-unitedqa}.
Consistent with previous work, our results also suggest that extra knowledge source is beneficial for ODQA.
Notably, we find that the verbalized knowledge is favored by the reader compared to the raw format (linearization), especially when the structured data size is comparable to text, leading to more pronounced improvements.
Overall, \ourmodel\space shows large improvements over text-only baselines and performs competitively with more complicated methods on both Natural Questions (NQ) \cite{kwiatkowski-etal-2019-natural} and WebQuestions (WebQ) \cite{berant-etal-2013-semantic}.
In particular, \ourmodel\space achieves new SOTA on NQ under the single-model open-book setting.\footnote{Data and code available at \url{https://github.com/Mayer123/UDT-QA}}

\section{Overview of \ourmodel}
\label{sec:approach}
\begin{figure*}
    \centering
    \includegraphics[scale=0.38]{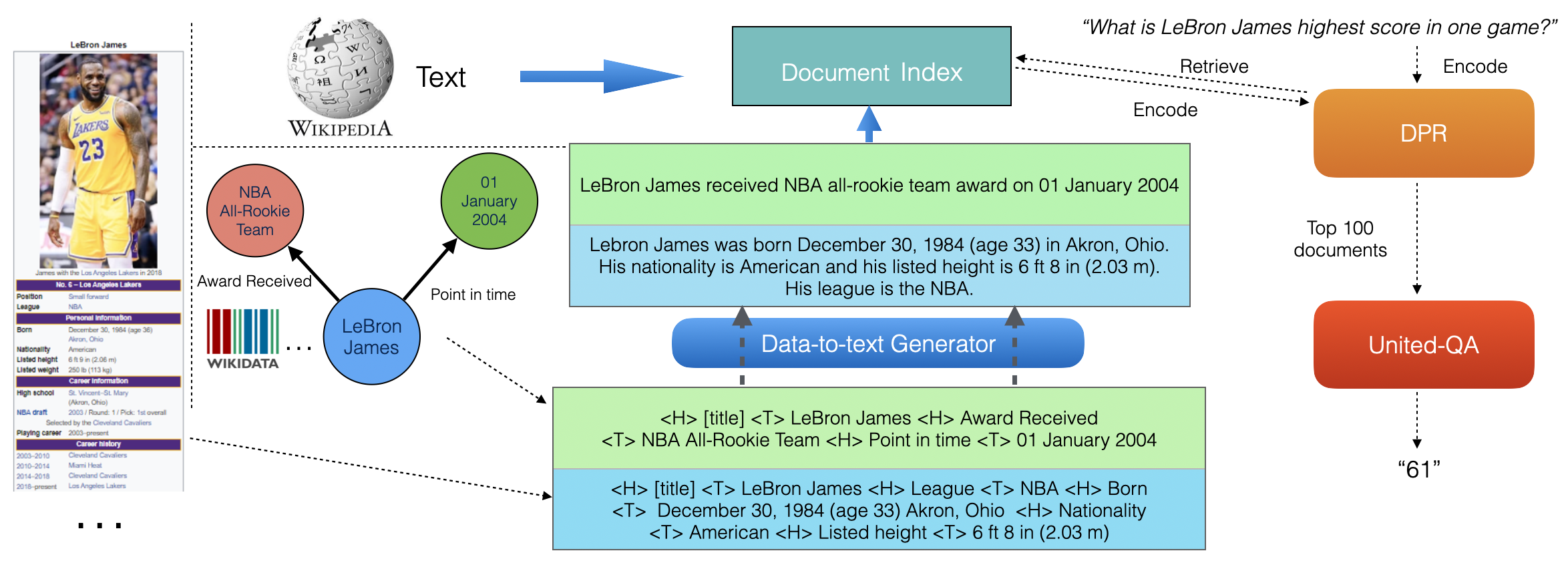}
    \caption{An overview of \ourmodel\space based on the \textit{verbalizer-retriever-reader} pipeline.}
    \label{fig:pipeline}
\end{figure*}

In this section, we present the overall pipeline of our \ourmodel\space framework for ODQA over data and text (\autoref{fig:pipeline}).
The major difference between our approach and the popular \textit{retriever-reader} ODQA systems \cite[\textit{inter alia}]{min2021neurips} is the use of a data-to-text verbalizer (\S\ref{sec:verbalizer}) for converting structured data into natural language text, \ie virtual documents, as the universal knowledge source. Here, we consider two types of structured knowledge (\S\ref{ssec:data_src}) --- tables and KB sub-graphs.
After verbalizing the structured knowledge, a subsequent pipeline consisting of a DPR retriever and a UnitedQA-E reader is used for answer inference.
Since the retriever and reader are not the main focus of this work, we only briefly describe them below.

The DPR retriever \cite{karpukhin-etal-2020-dense} is a bi-encoder model consisting of a question encoder and a context encoder, which is used for data and text retrieval.
Following previous work \cite{karpukhin-etal-2020-dense,oguz2020unikqa}, we use the uncased BERT-base \cite{devlin-etal-2019-bert} model as the encoder, where the \clstoken token representation is used as the document/question vector.
During training, positive and negative pairs of (question, context) are used to update the model.
For inference, the entire document index is encoded with context encoder and the encoded question vector is used to retrieve the top documents with highest dot-product scores.

The UnitedQA-E \cite{cheng-etal-2021-unitedqa} is an extractive reader based on ELECTRA \cite{clark2020electra} trained with enhanced objectives \cite{cheng-etal-2021-posterior,cheng-etal-2020-probabilistic} for answer inference. 
Here, a pair of a question and a support passage is jointly encoded into neural text representations.
These representations are used to compute scores of possible answer begin and end positions, which are then used to compute probabilities over possible answer spans. Finally, the answer string probabilities are computed based on the aggregation over all possible answer spans from the entire set of support passages.

\section{Verbalizer: Data-to-text Generation}
\label{sec:verbalizer}
Here, we formally describe the data-to-text model developed in this paper,
including the input format (\S\ref{ssec:dot_def}) 
and the adaptation for ODQA (\S\ref{ssec:dot_train}).
\subsection{Input Format}
\label{ssec:dot_def}

Given a structured data input $D$, the data-to-text generator $G$ aims to generate a natural language passage $P$ that faithfully describes the information presented in $D$.
In the literature, the structured data input can be in the form of a set of triples \cite{nan-etal-2021-dart}, a few highlighted cells from a table \cite{parikh-etal-2020-totto} or a full table \cite{chen-etal-2020-logical}.
Correspondingly, $P$ could a simple surface-form verbalization of $D$ (\eg when $D$ is a triple set) or a high-level summarization in case of a full table or a large KB graph.
Since we consider (noisy) tables/KB sub-graphs of arbitrary size in this paper, directly feeding the entire input into the generator is not feasible, likely incurring significant computation challenges.
Moreover, it is also desirable to maximize the information coverage of $P$ so that most relevant information in $D$ can be leveraged by the downstream QA retriever and reader.
Based on this, we verbalize both tables and KB graphs at a fine-grained level. 

In this work, we verbalize tables row by row, i.e. input each table row to $G$ individually, where each row is a set of cells $r=\{c_i\}_{i=1}^k$, and $k$ is the number of cells in the corresponding row.
Most relevant to our setting, recent work \cite{nan-etal-2021-dart} represents each cell in a triple.
To form such triples, they manually annotate the tree ontology of column headers and then create triples using table title, headers, cell value and header relations, \eg (\texttt{[TABLECONTEXT], [title], LeBron James}), (\texttt{LeBron James, League, NBA}) where \texttt{LeBron James} is the parent cell.
Although such triples with fine-grained ordering may help guide the generator,
directly applying such a generator to a target domain with no ontology annotation (our case) likely results in degradation.
To overcome this, we propose to convert the triple set to pairs, \eg
(\texttt{[title], LeBron James}),
(\texttt{League, NBA}).
We find such conversion has little impact on the intrinsic evaluation (\S\ref{sec:exp_verb}).
After all rows are verbalized, we assemble the text outputs back to form the verbalized table. 

For KB, we follow previous work \cite{agarwal-etal-2021-knowledge} and break the KB into small sub-graphs based on subject entity.
Here, each sub-graph contains one central entity and its neighbors.
Although this conversion would inevitably create undesirable artifacts (\eg hurdles for multi-hop reasoning across sub-graphs),
this preprocessing allows us to unify the input representations for both table and KB graphs, making it possible for a single verbalizer to convert structured knowledge into text format.
Specifically, we convert all KB sub-graphs into the same format as table cell sets above, where the subject entity is treated as the title and all the edges are represented using pairs in the form of (\texttt{relation}, \texttt{object}).
Then we verbalize each sub-graph with the generator $G$. Examples of input and output for table rows and KB sub-graphs are shown in \autoref{fig:pipeline}.

\subsection{Improved Data-to-Text Model Training} 
\label{ssec:dot_train}

A known problem in data-to-text generation is that the model tends to hallucinate or neglect information in the input \cite{wang-etal-2020-towards,agarwal-etal-2021-knowledge}.
Faithfulness and information coverage is especially important when we apply the verbalized output to knowledge-intensive downstream tasks like ODQA. 
To address this, we subsample training data \texttt{T} such that the instances are filtered out if they are likely to steer model towards missing information.
In particular, we compute ROUGE-1 \cite{lin-2004-rouge} scores between the input and target of training instances and filter out those whose scores are below a certain threshold.
We denote the filtered version as \texttt{T-F}.
Examples of the filtered instances can be found in \autoref{tab:data_to_text_exp}, as we discuss more in \autoref{sec:supp_data_to_text}, these instances may bias the model towards unwanted behaviors.

Another challenge we face is that most data-to-text training examples have succinct structured inputs.
In other words, the cells in the structured input are usually single words or short phrases with corresponding short target sentences as well.
In our case, a number of tables contain large cells with dozens of words.
Models trained with existing data likely have a hard time verbalizing such inputs faithfully.
To alleviate this domain-mismatch issue, we propose an iterative training set-up.
In the first iteration, we train a generator on \texttt{T-F}. Then we apply the generator to our data. We then find high quality verbalized outputs based on the ROUGE-1 score between the model inputs and model outputs, and sample instances with score higher than a threshold for the next-round training. We sample instances up to the same size of \texttt{T-F}, and denote this set as \texttt{ID-T} (examples shown in \autoref{tab:data_to_text_exp}). Finally, we mix the \texttt{ID-T} with \texttt{T-F} and train a second generator for verbalization. 

Following recent work \cite{nan-etal-2021-dart}, we use the pretrained T5-Large \cite{2020t5} model as our generator.
Given paired training examples consisting of a structured data input and a target sentence, we finetune the T5 model to maximize the log-likelihood of generating the corresponding target sentences. Here, we follow the same experimental setup as \cite{ribeiro2020investigating}. 

 \begin{table*}[hbt!]
\centering
\resizebox{\linewidth}{!}{
\begin{tabular}{lc|cccccc|c}
\toprule
 & & \multicolumn{6}{|c|}{Intrinsic Eval} & Extrinsic Eval\\
\midrule
 \textbf{Training Set} & \textbf{\# Examples} &  \textbf{BLEU} & \textbf{METEOR} & \textbf{TER} & \textbf{MoverScore} & \textbf{BERTScore} & \textbf{BLEURT} & \textbf{Ans Cov}\\
\midrule
DART \cite{nan-etal-2021-dart} & 62,659  & 50.66 & 0.40 & 0.43 & 0.54 & 0.95 & 0.44 & - \\
DART ours (\texttt{T}) & 62,628  & 51.05 & 0.40 & 0.43 & 0.54 & 0.95 & 0.43 & 95.4  \\
DART (\texttt{T-F}) & 55,115  & 51.04 & 0.41 & 0.43 & 0.54 & 0.95 & 0.43 & 96.0  \\
DART (\texttt{T-F + ID-T})  & 110,230  & 50.59 & 0.41 & 0.44 & 0.54 & 0.95 & 0.43 & \textbf{98.4} \\
\bottomrule
\end{tabular}
}
\caption{Intrinsic and extrinsic evaluations of verbalization approaches on DART test and \texttt{NQ-table-Q} (\S\ref{ssec:datasets}), respectively. ``Ans Cov'' refers to Answer coverage. All metrics are higher the better except for TER.}
\label{tab:dart_eval}
\end{table*}

\section{Experiment Setup}
\label{sec:exp_setup}
In this section, we describe the data used for experiments and sources of structured knowledge.

\subsection{Datasets}
\label{ssec:datasets}
In this paper, we use DART \cite{nan-etal-2021-dart} to train our verbalizer (data-to-text) and two ODQA datasets, NQ and WebQ, to train and evaluate our pipeline, with the same split as in \cite{lee-etal-2019-latent} provided by \cite{karpukhin-etal-2020-dense}. Below we provide a brief description of each dataset and refer readers to their papers for details.

\noindent \textbf{DART} is a data-to-text dataset containing pairs of (triple-set, sentences) collected from WebNLG \cite{gardent-etal-2017-webnlg}, E2E  \cite{novikova-etal-2017-e2e} and crowdsourcing based on tables found in WikiSQL \cite{zhong2017seq2sql} and WikiTableQuestions \cite{pasupat-liang-2015-compositional}. \\
\noindent \textbf{Natural Questions} contains questions mined from Google search queries and the answers are annotated in Wikipedia articles by crowd workers. \\
\noindent \textbf{WebQuestions} consists of questions from Google Suggest API and the answers are annotated as entities in Freebase. 

We collect {\bf knowledge-answerable questions} from NQ and WebQ in order to evaluate our verbalizer and construct the retrieval training data.
Specifically, we find questions in the original NQ training set that can be answered by a table. For each question, we search through tables in its associated HTML page to locate exact answer matches.
In total, we collected 14,164 triples of (question, answer, gold table) from NQ train and dev sets as \texttt{NQ-table-Q}. On WebQ, we find questions that can be answered by KB via expanding from question entities and search for their 1-hop neighbors. If an answer entity is matched, we keep this sub-graph. In total, we collected 2,397 triples of (question, answer, sub-graph) from WebQ train and dev set as \texttt{WebQ-KB-Q}.

\subsection{Structured Knowledge Sources}
\label{ssec:data_src}
In addition to regular Wikipedia text passages, we consider two types of structured knowledge --- tables from Wikipedia and KB graphs from Wikidata.

For tables from Wikipedia, we follow OTT-QA \cite{chen2021ottqa} with slight modifications.
\citet{chen2021ottqa} only consider tables in good format, \ie tables with no empty cell, multi-column or multi-row, and restrict the tables to have at most 20 rows or columns. Instead, we remove such constraints and keep everything with the \texttt{<table>} tag, resulting in a larger and noisier table set.
We denote this more realistic set of tables as \texttt{OTT-tables}. 

Note \citet{oguz2020unikqa} only consider tables from the original NQ HTMLs.
In addition to the size difference, \texttt{OTT-tables} are crawled from a more recent Wikipedia dump than the NQ version.
To study the impact of knowledge source size, we also process tables from the NQ HTML pages with the heuristic suggested by \cite{herzig-etal-2021-open} to de-duplicate tables and filter lengthy cells (>80 words).
We denote this set of tables as \texttt{NQ-tables}.
To avoid overlap, we remove tables from \texttt{OTT-tables} whose page title are in \texttt{NQ-tables} set.
In total, we have a \texttt{All-tables} set with 2.2M tables from \texttt{OTT-tables} and 210K tables from \texttt{NQ-tables}, respectively.

For KB graphs, we consider using the English Wikidata \cite{10.1145/2629489} as our
KB due to its broad coverage and high quality, noting its predecessor Freebase is no longer maintained despite its popularity in research.
In order to be comparable with recent work \cite{agarwal-etal-2021-knowledge}, we directly use their partitioned KB graphs from WikiData in our experiments, which is denoted as \texttt{WD-graphs}.

\section{Experiments: Data-to-Text}
\label{sec:exp_verb}
In this section, we evaluate our data-to-text model with both intrinsic and extrinsic metrics.
Since intrinsic metrics are probably less correlated with the downstream performance,
we use them only as a sanity check for generation quality and focus on using an extrinsic metric for selecting models.

\noindent {\bf Intrinsic Evaluation}:
Since our model is developed mainly on DART, we first conduct the intrinsic evaluation on the DART test set to measure the impact of our improved data-to-text methods, \ie 
data filtering and iterative training.
Following \cite{nan-etal-2021-dart}, we use the official evaluation metrics including BLEU, METEOR \cite{banerjee-lavie-2005-meteor}, TER, MoverScore \cite{zhao-etal-2019-moverscore}, BERTScore \cite{Zhang2020BERTScore:} and BLEURT \cite{sellam2020bleurt}.
\autoref{tab:dart_eval} summarizes different data-to-text models on DART test.
As we can see, the resulting model trained with our data conversion (row 2) performs on par with the model using the original format (row 1).
More interestingly, filtering short samples has almost no impact on the verbalizer performance (row 3).
Lastly, iterative training with additional target domain data (row 4) slightly hurts on BLEU and TER and achieves similar performances on other metrics.
Overall, our verbalizer with the proposed data conversion and improved training remains very effective on DART.

\noindent{\bf Extrinsic Evaluation}:
Since we are interested in applying verbalized knowledge for ODQA, the QA model is more likely to predict the correct answer only if the answer still exists after the verbalization. 
Therefore, we also evaluate each generator using a metric more related with the downstream task performance: \textbf{answer coverage}.
Specifically, we compute the answer coverage as the percentage of examples that the answer present in the raw structured knowledge is still preserved in the corresponding verbalized output.

First, we compute the answer coverage of different generators discussed in the previous section on \texttt{NQ-table-Q} where tables are known to contain question-triggering content. 
The scores are reported in the last column of \autoref{tab:dart_eval}.
Due to more lengthy tables in \texttt{NQ-table-Q}, data filtering improves the answer coverage as expected.
Moreover, model trained with our iterative training demonstrates substantial improvements in answer coverage, indicating that our approach is highly effective for converting tables into text. Examples for comparing different verbalizer outputs are shown in \autoref{tab:verbalizer_compare} in \autoref{sec:supp_data_to_text}. Later, we use this best generator to verbalize \texttt{All-tables}. We use beam search of size 10 and save all beams. To retain as much input information as possible, a re-ranking stage is carried out over these predictions based on the ROUGE-1 score between the model inputs and model outputs. The highest ranked prediction is then used as the final output. 

Lastly, we directly apply our best generator (DART T-F + ID-T) for verbalizing KB graphs.
To evaluate the performance, we compare our model with the recent method KELM-verbalizer \cite{agarwal-etal-2021-knowledge} using answer coverage on the set \texttt{WebQ-KB-Q} where KB sub-graphs are known to contain answer entities.
Although never tuned for KB graph inputs, our model achieves 99.6 on answer coverage, outperforming the KELM-verbalizer (97.8 on answer coverage) by a large margin.
This suggests that our data-to-text approach is highly effective for both tables and KB sub-graphs.

\section{Experiments: QA over Data and Text}
\label{sec:exp_qa}
Here we present our main experiments on ODQA over data and text.
For regular Wikipedia text, we use the same index containing 21M passages as in \cite{karpukhin-etal-2020-dense}.
To augment text, two settings are considered, \ie the \textit{single data} setting and the \textit{hybrid data} setting.


In the single data setting for NQ, we augment the text index with tables from the \texttt{All-tables} set (\S\ref{ssec:data_src}).
For comparison, we also experiment with the raw representations using a simple linearization of tables similar to \cite{oguz2020unikqa}.
In single data setting for WebQ, we consider combining text with KB graphs from \texttt{WD-graphs} in the single data setting.
Different from \cite{oguz2020unikqa} where a separate entity-linking based retriever is used for KB, we use a single model over the text index with either linearization of raw KB graphs or our verbalized KB graphs. 
Hence, in our case, both text and data (tables and KB graphs) can be handled by a unified retriever-reader pipeline. 
In the hybrid data setting for both NQ and WebQ, we use text, \texttt{All-tables} and \texttt{WD-graphs} for retrieval.
The statistics of our document index are shown in \autoref{tab:index} in \autoref{sec:index_size}.

We create additional retriever training data from \texttt{NQ-Table-Q} and \texttt{WebQ-KB-Q} in a similar fashion as in the text-only setting, so that DPR can better handle additional knowledge.
Following \cite{oguz2020unikqa}, we also use the iterative training set-up for retriever training. More training details can be found in \autoref{sec:details}. 

\begin{table}[t]
    \centering
    \small
    \begin{tabular}{l|c|c}
    \hline
\toprule
\multicolumn{1}{c|}{\bf Model} 
&  {\bf NQ} & {\bf WebQ} \\

\midrule
\multicolumn{3}{c}{\it Without Structured Knowledge} \\
\midrule
DPR \cite{karpukhin-etal-2020-dense}    
& 41.5 & 35.2 \\
UnitedQA \cite{cheng-etal-2021-unitedqa} 
& 51.8 &  48.0\\
\midrule
\multicolumn{3}{c}{\it With Structured Knowledge} \\
\midrule
KEALM \cite{agarwal-etal-2021-knowledge} 
& 41.5 & 43.9 \\
UnitK-QA \cite{oguz2020unikqa} 
&  54.1 & \textbf{57.8} \\
\ourmodel\space \textit{w/} Raw Single Data
& 54.7 & 51.4\\
\ourmodel\space \textit{w/} Verbalized Single Data
& \textbf{55.2} & 52.0\\
\ourmodel\space \textit{w/} Verbalized Hybrid Data
& 55.1 & 52.5\\

\bottomrule
    \end{tabular}
    \caption{End-to-end open-domain QA evaluation of \ourmodel\space in comparison to recent state-of-the-art models on the test sets of NQ and WebQ. Exact match scores are reported (highest scores shown in \textbf{bold}).}
    \label{tab:main_results}
\end{table}

To evaluate the effectiveness of our \ourmodel\space for ODQA, we first include recent state-of-the-art ODQA models using text as the only knowledge source, DPR and UnitedQA.
We also compare our \ourmodel\space with recent models using additional structured knowledge, KEALM and UnitK-QA.
Following the literature, we report the exact match (EM) score for evaluation.
The results are in \autoref{tab:main_results}.

As we can see, models with additional structured knowledge achieve better performance than text-only models. This indicates that both KB graphs and tables contain complementary knowledge which is either absent in text or harder to be reasoned over. 
For NQ, although we consider a significantly larger structured knowledge source which is likely to be more
challenging, all our models substantially outperform UnitK-QA.
As for WebQ, our model achieves competitive performance, although worse than UnitK-QA.
We attribute this gap to two possible reasons.
First, UnitK-QA uses a separate entity-linking based retriever for KBs which might lead to higher retrieval recall.
Second, since WebQ is fully based on FreeBase, using WikiData only in our models likely suffers from mismatch \cite{10.1145/2872427.2874809}.
Nevertheless, our verbalizer-based models achieve better performances than the corresponding raw format models on both datasets, indicating that the proposed verbalizer is highly effective for tables and KB graphs.

\begin{table}
\resizebox{\linewidth}{!}{
\begin{tabular}{lcccc}
\toprule
\textbf{Source} & \textbf{Format} & \textbf{R20} & \textbf{R100} & \textbf{EM}\\
\hline
text  & - & 80.8 & 86.1 & 49.6 \\
+\texttt{NQ-tables}  & raw & 85.2 & 90.1 & 51.1 \\
+\texttt{NQ-tables}  & V & 85.5 & 90.2 & 51.2\\
+\texttt{All-tables} & raw & 85.8 & \bf 90.7 & 52.1\\
+\texttt{All-tables} & V & \bf 86.0 & \bf 90.7 & \bf 52.5 \\
\midrule
text  & - & 78.9 & 82.3 & 52.6 \\
+\texttt{WD-graphs-WebQ}  & raw & \bf 83.4 & 86.1 & \bf 57.1 \\
+\texttt{WD-graphs-WebQ}  & V & \bf 83.4 & 85.0 & 55.7 \\
+\texttt{WD-graphs}  & raw & 82.8 & 86.1 & 54.3 \\
+\texttt{WD-graphs} & V & 82.8 & \bf 86.7 & 55.4\\
\bottomrule
\end{tabular}
}
\caption{Impact of document index size over separately trained retriever-reader models (Top for NQ and bottom for WebQ). All metrics are computed on the corresponding dev set. V stands for Verbalized here and on-wards.}
\label{tab:retrieval}
\end{table}

\section{Analysis}
\label{sec:analysis}
In this section, we present analyses over the impact of document index size,
the use of additional structured knowledge in a hot-swap setting,
comparison to a recent KB-only data-to-text approach in an end-to-end fashion,
and manual exam of the verbalized/raw tables for their impact on ODQA.

\noindent
\textbf{How does the size of document index affect retriever and reader performance?}
More knowledge is likely to have better coverage of relevant information.
On the other hand, larger and noisier index also increases the reasoning complexity. 
To understand the impact of the increased document index size,
we conduct experiments with a restricted setting where only relevant subset of knowledge to the corresponding dataset (a prior) is used for retrieval. Similar to \cite{oguz2020unikqa}, we experiment with the combined document index of text and \texttt{NQ-tables} for NQ. As for WebQ, we keep documents from \texttt{WD-graphs} that contain any of the question entity in 
WebQ to build \texttt{WD-graphs-WebQ}, and experiment with using text + \texttt{WD-graphs-WebQ}.
In addition to EM, we report R20 and R100, evaluating the retrieval accuracy of gold passages in the top-20 and top-100 documents, respectively.
The results are reported in \autoref{tab:retrieval}. 

\begin{table}
\centering
\resizebox{\linewidth}{!}{
\begin{tabular}{lcccc}
\toprule
\textbf{Source} & \textbf{Format} & \textbf{R20} & \textbf{R100} & \textbf{EM}\\
\midrule
\texttt{Text-only}  &     & 81.3 & 87.3 & 51.8 \\
\midrule
+\texttt{NQ-tables}  & raw & 83.9 & 90.3 & 51.7  \\
+\texttt{NQ-tables}  & V   & 84.3 & 90.4 & 52.5 \\
\midrule
+\texttt{All-tables} & raw & 84.0 & \bf 90.6 & 51.7 \\
+\texttt{All-tables} & V   & \bf 84.5 & \bf 90.6 & \bf 52.7 \\
\bottomrule
\end{tabular}
}
\caption{Hot-swap evaluation of raw vs verbalized table using a text-only retriever-reader model on NQ test.}
\label{tab:swap}
\end{table}

For NQ, in spite of being more challenging, we see that using \texttt{All-tables} yield substantial improvement in both recall and answer exact match compare to using \texttt{NQ-tables}.
This indicates that, with proper training, ODQA models are likely to benefit from enriched knowledge.
Although the larger raw form index brings in decent improvement (+1 EM) in terms of reader performance (+\texttt{All-tables} vs+\texttt{NQ-tables}), our verbalized knowledge is more friendly for answer reasoning leading to a more notable QA improvement (+1.3 EM).
Different from NQ, we observe that on WebQ the restricted setting with \texttt{WD-graphs-WebQ} achieves better results.
We hypothesize that this is likely due to the scale of WebQ dataset.
The small amount of WebQ training makes the retriever insufficient to handle large-scale document index.
We leave the verification of this hypothesis for future work. 

\noindent
\textbf{Does a text-only retriever-reader model benefit more from verbalized knowledge compare to raw format (hot-swap)?}
Since both retriever and reader are based on pretrained language models, we hypothesize that they would probably benefit more from the verbalized knowledge due to its similar style as text. This can be particularly useful for a hot-swap setting where both retriever and reader have only seen textual knowledge during training. 
To verify that verbalized knowledge is more amenable, we carry out a hot-swap experiment here.
Specifically, we directly use a DPR model trained on NQ text-only data for additionally indexing both \texttt{NQ-tables} and \texttt{All-tables}.
Then, the inference retrieval is performed on the augmented document index for an input question, and a text-only United-QA-E reader trained on NQ is applied for answer inference afterwards.
The results are summarized in \autoref{tab:swap}.
Similar to the previous fully fine-tuned settings, we see that additional knowledge still provide substantial improvements for text-only retriever using
either raw or verbalized knowledge.
However, the improvement in recall is not reflected in the later reader performance for the raw format, whereas the hot-swap answer inference performance is notably improved with verbalized knowledge.
This observation further validates our hypothesis that verbalized knowledge is more beneficial, especially for reader.  

\begin{table}
\centering

\begin{tabular}{lccc}
\toprule
\textbf{Source} & \textbf{R20} & \textbf{R100} & \textbf{EM}\\
\hline
\texttt{KELM}   & 78.2  & 85.3 &  51.5\\
\texttt{WD-graphs} (Ours)    & \bf78.5 & \bf 85.5 & \bf 52.0 \\
\bottomrule
\end{tabular}
\caption{Comparison of verbalized knowledge from our verbalizer and KELM for retriever and reader on WebQ test.
Dev results can be found in \autoref{tab:kelm-dev} in \autoref{sec:supp_comp_our_vs_kelm}. }
\label{tab:kelm}
\end{table}

\begin{table*}[h!]
\small
\centering
\resizebox{\linewidth}{!}{
\begin{tabular}{lll}
\toprule
 \textbf{Q\&A}& \textbf{V table} & \textbf{Raw table} \\
\hline
& \textbf{TITLE:} List of Star Wars: The Clone Wars episodes &  \\
\textbf{Q:} star wars & .... the theatrical film: "the new padawan" "castle of & | no. in series, season, no. in season, title |\\
the clone wars & deception" "castle of doom" "castle of  salvation" is no. & .... | 3-6, empty, empty, theatrical film: "the \\
season 3  &  3-6 in the series of star wars: the clone wars episodes.  & new padawan"  "castle of deception" "castle \\
episode 1  & \textbf{"clone cadets" in season 3 of star wars: the clone } & of doom" "castle of salvation" | \textbf{ 7, 3, 1,} \\
\textbf{A:} Clone Cadets & \textbf{wars is number 1 in season and number 7 in series.}  & \textbf{"clone cadets"} | 8, 3, empty, "supply lines" | \\
& "supply lines" is episode 8 in series and 3 in season of & .... \\
& star wars: the clone wars game .... & \\
\midrule
&\textbf{TITLE:} Mount Ruapehu & \\
\textbf{Q:} when was & .... mount ruapehu is a stratovolcano mountain with & | empty, empty, empty, elevation, prominence, \\
the last time & an age of 200,000 years. \textbf{the last eruption was 25} & listing, coordinates, empty, translation, empty, \\
mount ruapehu & \textbf{september 2007} and the volcanic arc/belt is taupo & empty, empty, age of rock, mountain type,  \\
erupted & volcanic zone. mount ruapehu was first ascent in & volcanic arc/belt, \textbf{last eruption}, empty, first \\
\textbf{A:} 25 September & 1879 by g. beetham and j. p.  maxwell. the easiest & ascent, easiest route | .... ~200,000 years, strato- \\
2007 & route to climb mount ruapehu is hike. & volcano, taupo volcanic zone, \textbf{25 september} \\
&  & \textbf{2007}, climbing, 1879 .... |\\
\midrule
& \textbf{TITLE:} List of National Football League career & rushing yards leaders \\
\textbf{Q:} who has & \textbf{emmitt smith} of the dallas cowboys (1990-2002) & | rank, player, team(s) by season, carries, \\
the most & and arizona cardinals (2003-2004) \textbf{was the first} & yards, average | \textbf{1, emmitt smith}, dallas \\
yards per carry & \textbf{player on the national football league career} & cowboys   (1990-2002) arizona cardinals \\
in nfl history  & \textbf{rushing yards leaders list}. walter payton of the & (2003-2004),  4,409, 18,355, 4.2 | 2, walter \\
\textbf{A:} Emmitt Smith & chicago bears (1975-1987) ranked second .... & payton, chicago bears ....\\
\midrule
& \textbf{TITLE:} List of European countries by population & \\
\textbf{Q:} which country & .... \textbf{vatican city ranks 50 on the list of european} & | rank, country, current population, \% of \\
has the smallest & \textbf{countries by population with 1,000 current } & population, average relative annual growth(\%), \\
population in & \textbf{population} and 0.0 \% of population. the list of & average absolute annual growth, estimated \\
europe & european countries by population has 0.0 average & doubling time(years), official figure, date of \\ 
\textbf{A:} Vatican & relative annual growth(\%) and 0 average absolute & last figure, regional  grouping, source | 1 ....\\
 City & annual growth. the  source is official estimate and & 49 .... | \textbf{50, vatican city, 1,000}, 0.0, 0.0, 0, -, 0, \\
& the date of last figure is 2012. The total population ....  & 2012, empty, official estimate | empty, total, ....  \\
\bottomrule
\end{tabular}
}
\caption{Examples of tables/chunks retrieved by our model given the question, where the evidence is bolded. In raw table, | is the row separator and empty is the filler token used by our table parsing heuristic (to make the table in good shape)}
\label{tab:tab_v_raw}
\end{table*}

\noindent
\textbf{How does the proposed verbalizer compare to recent data-to-text models?}
Lastly, we compare our verbalizer with the recently proposed data-to-text generator for converting KB graphs only, KELM \cite{agarwal-etal-2021-knowledge}.
Since both KELM generator and our verbalizer are based on the same partitioned Wikidata, this evaluation can fully reflect their corresponding generation impacts on ODQA in an end-to-end fashion.
Here, we evaluate using our verbalized \texttt{WD-graphs} and the KELM corpus as additional knowledge on WebQ.
In particular, we follow the same procedure to train and evaluate our retriever and reader except that we swap the \texttt{WD-graphs} with KELM corpus in data construction and retrieval.
Both retriever and reader performances are reported in \autoref{tab:kelm}.
Note that the KELM data-to-text model is customized solely for converting KB graphs and trained with a much larger dataset (about 8M training instances),
whereas our verbalizer is applicable to both tables and KB graphs with a smaller training data (only 110K instances).
Nevertheless, consistent with its better extrinsic performance (\S\ref{sec:exp_verb}), our verbalizer again outperforms the KELM generator in both retrieval and reading,
which provides further support for the effectiveness of our approach as a unified interface for ODQA over data and text. \\
\noindent\textbf{What is the impact of verbalized/raw table on ODQA?}
We manually analyze examples of verbalized and raw tables and the details of annotation can be found in \autoref{sec:error}.
We showcase the examples of verbalized tables and their raw counterpart in \autoref{tab:tab_v_raw} and discussion their effect on our \ourmodel\space system. 
We identify 2 common patterns where raw tables are inferior to verbalized tables, as shown in the first 2 rows of \autoref{tab:tab_v_raw}. In the first example, \textit{the concatenated numbers in the raw table can be hard to interpret}, and we have to carefully align the row with the header, which is very far away. In the second example, \textit{the raw infobox can be in ill-format and very long}, making it hard to understand. On the other hand, the verbalized row clearly states the answer evidence by connecting the information in the headers with cell values, making it straightforward to find the answer.

At the same time, we also notice the limitation of verbalized tables: table structure loss. We found that raw tables are better at answering ranking questions, as the examples shown in row 3\&4 of \autoref{tab:tab_v_raw}. When asked about the top or bottom ranked subject, the model can directly look for evidence from the starting or the end of the table. On the other hand, when the table is verbalized, the model can not rely on such shortcuts because the boundary of rows is not clear and \textit{the original structure of the tables are lost}. This also suggests a possible direction for future work: to better incorporate the table structure information in verbalization. 
\section{Related Work}
\label{sec:related_work}

\noindent
 
\noindent \textbf{Data-to-Text} Generating text from structured data has been a popular task in NLP. Many dataset have been proposed for this task such as Wikibio \cite{lebret-etal-2016-neural}, Rotowire \cite{wiseman-etal-2017-challenges},  WebNLG \cite{gardent-etal-2017-webnlg} and E2E \cite{novikova-etal-2017-e2e}, where each dataset focuses on a particular domain. More recently, large-scale datasets that contains open-domain examples have been proposed including DART \cite{nan-etal-2021-dart}, TOTTO \cite{parikh-etal-2020-totto}, WikiTableT \cite{chen-etal-2021-wikitablet} and GenWiki \cite{jin-etal-2020-genwiki}. On the modeling side, finetuning the pretrained models typically achieves promising performance \cite{ribeiro2020investigating}. \citet{wang-etal-2020-towards} propose customized loss functions to reduce model hallucination during generation.
Muti-task learning is used to improve model's robustness towards input variations \cite{hoyle-etal-2021-promoting}.
\citet{chen-etal-2020-kgpt} introduce a generalized format and a pretrained model that can generate text from both table rows and knowledge graphs.
Most previous work on data-to-text generation have only conducted internal evaluation, using typical generation metrics such as BLEU and ROUGE, hence the data-to-text is considered the target task.
In this paper, we argue that different training strategies and evaluation metrics should be adapted when applying data-to-text models to downstream tasks, i.e. ODQA.
Related to our work, \citet{agarwal-etal-2021-knowledge} convert the entire Wikidata to natural language using a finetuned T5 model \cite{2020t5}. 
In this work, we generalize the data-to-text approach for verbalizing both tables and KB graphs  in a unified fashion and study the verbalized knowledge on ODQA.

\noindent
\textbf{QA with Data and Text}
\noindent As the knowledge required to answer the questions may not be available in textual corpus, previous studies have sought to incorporate knowledge from difference sources such as tables and knowledge bases. \citet{min2019knowledge} use Wikidata to expand seed passages found by the retriever and enhance encoded passage representations in the reader.
\citet{li-etal-2021-dual} propose a hybrid framework that takes both text and tables as inputs to produce answers and SQL queries. 
Recently, \citet{chen2021ottqa} develop the OTT-QA dataset containing questions that require joint reasoning over both tables and text, where the tables and text come from entire Wikipedia.
There is also a line of work that studies model architectures for tables specifically or joint encoding of tables and text \cite{yin-etal-2020-tabert,herzig-etal-2020-tapas,zayats-etal-2021-representations, glass-etal-2021-capturing}.
However, their focus is not on open-domain QA tasks.
Most similar to our work is \cite{oguz2020unikqa}, where they use both tables and Wikidata/Freebase knowledge graph along with Wikipedia text for ODQA.
However, they simply linearized structured data without using any verbalizer, thus may suffer from sub-optimal input representation. Also, their tables are only mined from original NQ HTMLs, \ie a constrained setting.
In contrast, we consider tables from full Wikipedia which is a much larger set.
Additionally, separate retrieval models are used for tables and KB in \cite{oguz2020unikqa}
whereas we develop a unified model over text and data.

\section{Conclusion}
\label{sec:conclusion}
In this paper, we demonstrated that a unified \textit{verbalizer-retriever-reader} framework, \ourmodel, for open-domain QA over data and text.
We proposed a novel data-to-text paradigm that can largely improve the verbalization effectiveness for downstream knowledge-intensive applications, \ie open-domain QA, when attaining good intrinsic performances.
With the verbalized knowledge, we achieved a new state-of-the-art result for NQ.
Remarkably, we showed that simply augmenting the text index with the verbalized knowledge improve the performance without retraining the model.   

In addition to our method, there are many recently proposed approaches for open-domain QA that are orthogonal. For example, language models specifically optimized for dense retrieval \cite{gao2021unsupervised}, pretraining on large-scale QA data \cite{ouz2021domainmatched} and hybrid system that consists of retriever, reranker, extractive reader and generative reader \cite{fajcik2021r2d2}. Incorporating those methods may further improve the performance for open-domain QA, and we leave that exploration for future work. 
Lastly, instead of only considering a \textit{sanitized} collection of knowledge sources, it is an interesting future direction to scale up the knowledge to web-scale \cite{webgpt,webkilt}.

\section*{Acknowledgements}
We would like to thank Ruohong Zhang for helpful discussions and anonymous reviewers for their valuable suggestions on this paper.  

\bibliography{ref,qa}
\bibliographystyle{acl_natbib}
\clearpage

\appendix
\section{Document Index Statistics}
To be consistent with text passages, we also cut tables and KB sub-graphs (raw or verbalized) into chunks that has about 100 words. Hence the verbalized knowledge will have larger index size than raw format (see \autoref{tab:index}). 

\label{sec:index_size}
\begin{table}[t!]
\centering

\begin{tabular}{lcc}
\toprule
\textbf{Source} & \textbf{Raw} & \textbf{Verbalized} \\
\hline
Text & 21M  & -  \\
\texttt{OTT-tables} & 4.0M & 6.3M  \\
\texttt{NQ-tables} & 446K & 572K  \\
\texttt{WD-graphs} & 5.7M & 5.8M  \\
\bottomrule
\end{tabular}
\caption{Statistics of Document Index}
\label{tab:index}
\end{table}

\section{Training Details}
\label{sec:details}
To train the retriever to better handle knowledge from tables and KB, we create additional training data from \texttt{NQ-Table-Q} and \texttt{WebQ-KB-Q}. 
Given a (question, answer, gold table) from \texttt{NQ-Table-Q}, we create a positive passage by concatenating rows containing the answer. Then we randomly sample and concatenate other rows in the table if the passage has less than 100 words. To find negative passages for training, we build a index consists of all the tables and use BM25 to retrieve relevant tables. Ones that do not contain the answer are considered as negative tables. Then we sample rows from the table to build negative passages. For the raw tables, the process is the same except that we also concatenate headers in the beginning to build positive and negative passages. We combine NQ training data with this set to train DPR.

For \texttt{WebQ-KB-Q}, we use the verbalized gold sub-graphs as positive passages. For the raw format, this is replaced by flattening the gold sub-graph. Then we build an index with all documents in \texttt{WD-graphs} and the top ranked documents by BM25 that do not contain the answer are treated as negatives. Here the documents refer to concatenated triples set for raw setting and sentences produced by the generator in verbalized setting. Additionally, we search through answer entities and their neighbors in the graph to find documents that has word overlap with the question. Then we build training instances in a similar fashion. 

As pointed by previous work \cite{oguz2020unikqa}, mining harder negative passages using DPR and iterative training leads to better performance. We also adopted this approach in our experiments. After the first DPR is trained, we used it to retrieve passages from a joint index of text+\texttt{structured knowledge}. Then the negative passages are paired with the positive passages from the first round to build new sets of training data. Then we train a second DPR using the iteration1 data combined with the new training sets. 

For retriver training, we follow the experiment set-up as specified by \cite{karpukhin-etal-2020-dense}. Specifically, we use the Adam optimizer and a per-gpu batch size of 32 for NQ and 24 for WebQ, respectively.
All trainings are done with a fixed learning rate of $2e-5$ and 40 epochs using 8 V100 GPUs.
We select the best model based on the retrieval accuracy on the corresponding dev set.

For reader training, we follow the experiment set-up as described in \cite{cheng-etal-2021-unitedqa}. Specifically, we use the Adam optimizer and a batch size of 16 for NQ and 8 for WebQ, respectively. We use 16 and 8 V100 GPUs for NQ and WebQ respectively.
We select the learning rate in $\{3e-5, 5e-5\}$ and number of training epochs in $\{6, 8\}$.
The best model is selected based on EM on the corresponding dev set. All of our reported results are obtained from a single run.

Regarding the number of parameters in the model, our verbalizer is based on T5-large, which has 770M parameters. Our retriever is a bi-encoder model based on bert-base, which has 220M parameters. Our reader model is based on ELECTRA-large, which has 330M parameters. 

\begin{table}
\centering

\resizebox{\linewidth}{!}{
\begin{tabular}{lcccc}
\toprule
\textbf{Source} & \textbf{Format} & \textbf{R20} & \textbf{R100} & \textbf{EM}\\
\hline
text  & - & 81.3 & 87.3 & 51.8 \\
+\texttt{NQ-tables}  & raw & 86.0 & 91.2 & 54.8 \\
+\texttt{NQ-tables}  & V & 86.2 & 91.0 & 54.2\\
+\texttt{All-tables} & raw & 86.9 & \bf 91.9 & 54.7\\
+\texttt{All-tables} & V & \bf 87.0 & 91.7 & \bf 55.2 \\
\midrule
text  & - & 73.2 & 81.4 & 48.0\\
+\texttt{WD-graphs-WebQ}  & raw & \bf 80.2 & \bf 85.8 & 51.5 \\
+\texttt{WD-graphs-WebQ}  & V & 79.7 & 85.3 & \bf 52.6 \\
+\texttt{WD-graphs}  & raw & 78.8 & 85.1 & 51.4 \\
+\texttt{WD-graphs} & V & 78.5 & 85.5 & 52.0 \\
\bottomrule
\end{tabular}
}
\caption{Impact of document index size over separately trained retriever-reader models (Top for NQ and bottom for WebQ). All metrics are computed on the corresponding test set.}
\label{tab:retrieval-test}
\end{table}

\section{Impact of Document Index Size}
We report the test set results of models trained with different document index in \autoref{tab:retrieval-test} (corresponding to \autoref{tab:retrieval}). Overall, we observe similar trends. For NQ, the model benefits more from a larger document index while for WebQ the restricted setting yield better performance.  

\label{sec:supp_impact_kindex}

\section{Comparison betweeh Our Verbalizer and KELM-verbalizer}
We report the dev set results of WebQ models trained with our verbalized \texttt{WD-graphs} in comparison with KELM in \autoref{tab:kelm-dev} (corresponding to \autoref{tab:kelm}). 
\label{sec:supp_comp_our_vs_kelm}
\begin{table}
\begin{tabular}{lccc}
\toprule
\textbf{Source} & \textbf{R20} & \textbf{R100} & \textbf{EM}\\
\hline
\texttt{KELM}   & \bf 83.1  & \bf 86.7 &  55.1\\
\texttt{WD-graphs} (Ours)    & 82.8 & \bf 86.7 & \bf 55.4 \\
\bottomrule
\end{tabular}
\caption{Dev set results of models trained on WebQ with verbalized WD-graph and KELM}
\label{tab:kelm-dev}
\end{table}

\section{Case Study on Raw vs Verbalized Tables}
\label{sec:error}
\begin{table}
\centering
\begin{tabular}{lcc}
\toprule
 & \textbf{V-correct} & \textbf{V-error} \\
\hline
\textbf{Raw-correct}   & 1750  & 223 \\
\textbf{Raw-error}   & 242 & 1395 \\
\bottomrule
\end{tabular}
\caption{Error matrix of \ourmodel\space trained with text+\texttt{All-tables} in raw and verbalized format}
\label{tab:tab_case}
\end{table}
For manual analysis of verbalized and raw tables, we start by computing the error matrix of the NQ models trained with text+\texttt{All-tables} in both format, as shown in \autoref{tab:tab_case}. We then manually annotated 100 examples where only 1 format of knowledge successfully answered the question (50 for each format), and we select examples where at least 1 table chunk is marked as positive by the retriever. Out of 50 examples where verbalized tables contain the answer span, 40 of them are true positives that provide direct evidence to the questions. In 35 out of 40 questions, the retriever for the raw model actually find the same table/chunks that provide the answer. However, the model failed to extract answer for those cases and we think it's mainly because the raw format of the noisy tables can be hard for the model to reason over, as discussed in \autoref{sec:analysis}.

We then looked at the other group of 50 questions (raw format). 37 of them are true positives that contain direct evidence. Then in 30 out of 37 questions, the verbalized retriever is able to find the corresponding verbalized table/chunks that also contain the answer. The remaining cases are all due to retriever failed to find the true positive table chunks. In these 30 cases, the most noticeable pattern is that the model is able to leverage structural shortcut to arrive at the answer, suggesting the limitation of verbalized tables.

\section{Data-to-text Examples}
\label{sec:supp_data_to_text}
In the top half of \autoref{tab:data_to_text_exp} we show examples from DART that are filtered out by our method, i.e. low ROUGE scores between input and target. In the first example, information from 2 cells are completely omitted from the target. The model may learn to omit information from this kind of examples, which is problematic when we consider QA as our downstream task. Our filtering method is also able to prune noisy examples, as shown in row 2\&3, where there is little correspondence between input and target. In row 4, we show an example where the target contains the information not exist in the input. This kind of examples may teach the model to hallucinate which is also an unwanted behavior, hence they are also filtered out. 

In the bottom half of \autoref{tab:data_to_text_exp} we show examples from \texttt{ID-T} set, i.e. good quality input-output pairs produced by the verbalizer trained on \texttt{T-F} set, when applied to our table sets. These examples often have longer inputs and/or larger table cells and the information coverage rate in target is very high. By combining \texttt{ID-T} set with \texttt{T-F} set to train a new verbalizer, the model can more effectively learn to generate faithful description of the input. 

Finally, in \autoref{tab:verbalizer_compare} we show examples of question-answer pairs along with their associated gold tables(rows). For each raw input, we show its corresponding output generated by the verbalizer trained on \texttt{T} and the verbalizer trained on \texttt{T-F+ID-T}. We can see that the direction evidence to the answer is present in one of the raw table cells, however in all 3 cases the verbalizer(\texttt{T}) failed to generate that piece of information. On the other hand, the verbalizer(\texttt{T-F+ID-T}) faithfully described all information presented in the input, showing the effectiveness of our proposed method. 

\begin{table*}[h!]
\small
\centering
\resizebox{\linewidth}{!}{
\begin{tabular}{ll}
\toprule
Input & <H> [title] <T> Conde McCullough <H> bridge name <T> oregon city bridge <H> carries <T> oregon route 43 \\
& <H> \textbf{year completed} <T> \textbf{1922} <H> \textbf{total length} <T> \textbf{745 feet 227 m} <H> location <T> oregon city oregon \\
Target & conde mccullough helped build the oregon city bridge, which is part of oregon rt. 43. \\
\hline
Input & <H> [title] <T> \textbf{Yury Aleksandrovich Golovkin} <H> russian state <T> russian empire \\
 & <H> \textbf{date from} <T> \textbf{9 november 1818} <H> \textbf{date until} <T> \textbf{16 september 1822}  \\
 Target & the \textit{first ambassador of russia to austria} was \textit{prince dmitry mikhailovich golitsyn}. \\
\hline
Input & <H> [title] <T> 43 <H> \textbf{competition} <T> \textbf{afc champions league} \\
Target & the club was \textit{officially founded as lucky-goldstar football club in 1983, by the lucky-goldstar group.} \\
 & \textit{fc seoul have won six league titles, two fa cups, two league cups and one super cup.} \\
 \hline 
 Input & <H> [title] <T> Ana Maria Beldana Ferriera <H> hometown <T> santiago de los caballeros \\
 Target & ana maría beldaña ferriera, from santiago de los caballeros, was a \textit{contestant in the miss dominican republic 2002 contest.}\\
\hline
\hline
Input & <H> [title] <T> Meet the Fockers <H> edited by <T> jon poll lee haxall alan baumgarten <H> production company  \\
& <T> tribeca productions everyman pictures <H> distributed by <T> universal pictures (north america) \\
& dreamworks pictures  (international)  \\
Target & meet the fockers was edited by jon poll, lee haxall, alan baumgarten and distributed by universal pictures (north america) \\
& dreamworks pictures (international). the production company was tribeca productions. \\
\hline 
Input & <H> [title] <T> Lamar Hunt U.S. Open Cup <H> season <T> 2010 <H> player <T> paulo jr. nate jaqua \\
 & <H> team <T> miami fc seattle sounders fc <H> goals <T> 5  \\
Target & paulo jr. nate jaqua scored 5 goals for miami fc seattle sounders fc in the 2010 lamar hunt u.s. open cup. \\

\bottomrule
\end{tabular}
}
\caption{Top: examples from DART that are filtered out by our method, the \textbf{bold} cells are omitted information from target, and \textit{italic text} from target are likely to bias the model towards hallucination. Bottom: examples from (\texttt{ID-T}), which is generated by our 1st iteration verbalizer}
\label{tab:data_to_text_exp}
\end{table*}

\begin{table*}[h!]
\small
\centering
\resizebox{\linewidth}{!}{
\begin{tabular}{ll}
\toprule
Question & how many episodes in season 7 walking dead  \\
Answer & 16 \\
\hline
Input & <H> [title] <T> The Walking Dead (season 7) <H> country of origin <T> united states <H> \textbf{no. of episodes}  \\
 & <T> \textbf{16} <H> the walking dead (season 7) <T> release <H> original network <T> amc \\
Verbalizer (\texttt{T}) & the original network for the walking dead (season 7) is amc. the country of origin for the walking dead \\
& (season 7) is united states. \\
Verbalizer (\texttt{T-F+ID-T}) & the original network of the walking dead (season 7) is amc and the country of origin is united states. \\
& \textbf{the walking dead (season 7) has 16 episodes.} \\
\hline
\hline
Question & when did nigeria adopt the presidential system of government \\
Answer & 1963 \\
\hline
Input & <H> [title] <T> Federal government of Nigeria <H> federal government of nigeria <T> coat of arms of nigeria \\
& <H> \textbf{formation} <T> \textbf{1963; 55 years ago} <H> founding document <T> constitution of nigeria \\ 
Verbalizer (\texttt{T}) & the constitution of nigeria is the founding document of the federal government of nigeria which was formed 55 \\
& years ago. the federal government of nigeria has the coat of arms of nigeria. \\
Verbalizer (\texttt{T-F+ID-T}) & the constitution of nigeria is the founding document of the federal government of nigeria which was \\
& formed \textbf{in 1963; 55 years ago}. the federal government of nigeria has the coat of arms of nigeria. \\
\hline
\hline
Question & what year did they stop making the saturn vue \\
Answer & 2009 \\
\hline
Input & <H> [title] <T> Saturn Vue <H> saturn vue <T> overview <H> manufacturer <T> saturn corporation \\
& (2002-2007) opel (general motors) (2008-2010) <H> \textbf{production} <T> \textbf{2001–2009} <H> model years \\
& <T> 2002–2010 <H> saturn vue <T> body and chassis \\
Verbalizer (\texttt{T}) & saturn vue's body and chassis were manufactured by saturn corporation (2002-2007) and opel (general \\
& motors) (2008-2010) during the model years 2002–2010. \\
Verbalizer (\texttt{T-F+ID-T}) & saturn corporation (2002-2007) opel (general motors) (2008-2010) \textbf{manufactured the saturn vue from} \\
& \textbf{2001–2009} and model years 2002–2010. the saturn vue has a body and chassis. \\
\bottomrule
\end{tabular}
}
\caption{Examples of verbalized table(rows) generated by different verbalizer, where the direct evidences to the answer are marked in \textbf{bold}}
\label{tab:verbalizer_compare}
\end{table*}

\section{License}
We list the License of the software and data used in this paper below:
\begin{itemize}
    \item DPR: CC-BY-NC 4.0 License
    \item DART: MIT License 
    \item KELM: CC BY-SA 2.0 license 
    \item OTT-QA: MIT License 
\end{itemize}
\setcounter{table}{0}
\renewcommand{\thetable}{A\arabic{table}}

\setcounter{figure}{0}
\renewcommand{\thefigure}{A\arabic{figure}}

\end{document}